%% file: main.tex
\documentclass[sigconf,nonacm]{acmart}

\AtBeginDocument{%
  \providecommand\BibTeX{{%
    \normalfont B\kern-0.5em{\scshape i\kern-0.25em b}\kern-0.8em\TeX}}}
    
\setcopyright{acmcopyright}
\copyrightyear{2018}
\acmYear{2018}
\acmDOI{10.1145/1122445.1122456}

\acmConference[SIGIR '21]{Woodstock '18: ACM Symposium on Neural
  Gaze Detection}{June 03--05, 2018}{Woodstock, NY}
\acmBooktitle{Woodstock '18: ACM Symposium on Neural Gaze Detection,
  April 19---23, 2021, Ljubljana, Slovenia}
\acmPrice{15.00}
\acmISBN{978-1-4503-XXXX-X/18/06}
    
\author[] 
{Aym\'e Arango, Jorge P\'erez and Barbara Poblete}

\usepackage{setspace}  
\usepackage[utf8]{inputenc}
\usepackage{graphicx}
\usepackage{subcaption}
\usepackage{multirow}
\usepackage{graphicx}
\usepackage{amsmath}
\usepackage{graphics}
\usepackage{booktabs}
\usepackage{graphicx}
\usepackage{hyperref}
\DeclareMathOperator*{\mean}{mean}
\usepackage{graphicx}
\usepackage{pdfpages}

\begin{document}

%\title{A cross-lingual approximation to the multicultural problem of hate speech.}
\title{Cross-lingual hate speech detection\\based on multilingual domain-specific word embeddings}

\begin{abstract}
%%% Context
Automatic hate speech detection in online social networks is an important open problem in Natural Language Processing (NLP).
Hate speech is a multidimensional issue, strongly dependant on language and cultural factors.
%
%This makes difficult transferring knowledge from other domains to the hate speech detection problem.
%%% Need
Despite its relevance, research on this topic has been almost exclusively devoted to English, with limited coverage of other languages. 
Most supervised learning resources, such as labeled datasets and NLP tools, have been created for this same language. 
Considering that a large portion of users worldwide speak in languages other than English, there is an important need for creating efficient approaches for multilingual hate speech detection.
%
%Given the current scenario it is important that such approaches can leverage existing resources in their learning process.
%
%%% Objective/Goals
In this work we propose to address the problem of multilingual hate speech detection from the perspective of transfer learning.
Our goal is to determine if knowledge from one particular language can be used to classify other language, and to determine effective ways to achieve this.
We propose a hate specific data representation (i.e., hate speech word embeddings) and evaluate its effectiveness against general-purpose universal representations most of which, unlike our proposed model, have been trained on massive amounts of data.
%
%We performed an extensive comparative evaluation, using  different machine learning models on English, Spanish and Italian datasets.
%
We focus on a cross-lingual setting, in which one needs to classify hate speech in one language without having access to any labeled data for that language.
We show that the use of our simple yet specific multilingual hate representations improves classification results compared to other data representations. 
%, specially for underrepresented languages.
% 
We explain this with a qualitative analysis showing that our specific representation is able to capture some common patterns in how hate speech presents itself in different languages.
%Overall, our findings indicate that there are common patterns in how hate speech is exhibited across different languages and that we are able to capture some of these relationships through our proposed representation.
%
We expect that our proposal and results can be a first step towards identifying cross-lingual hate patterns specially in low-resource languages.

Our proposal constitutes, to the best of our knowledge, the first attempt for constructing multilingual specific-task representations. Despite its simplicity, our model outperformed the previous approaches for most of the experimental setups. 
The best path to take for solving the problem of cross-lingual hate-speech detection is still unknown, and our findings can orient future solutions toward the use of domain-specific representations.

%Thus, allowing us to identify new hate speech expressions and to optimize existing learning resources.
\end{abstract}
% Moreover, we identified some relations between different language terms that could help us to understand the complex phenomenon of hate.
% In addition, the analysis of cross-lingual hate speech classification can provide us with valuable insight about this type of expressions.
%
%For example, if there are language-independent patterns in hate speech across different cultures and languages, and if these patterns can be exploited to discover emergent hate subjects?
%
% In addition, our goal is to capture word context in a hate-specific scenario. 



\maketitle

\section{Introduction}
%%% CONTEXT AND MOTIVATION (social media and hate speech) %%%
%
Online social media platforms have become an important means of interaction among millions of users worldwide.
Timely information, including news and opinions, as well as political campaigns and other organized communications take place in this online environment.
However, along with many useful exchanges, there is also the manifestation of certain communication disorders such as {\em fake news} and {\em hate speech} which can produce harmful side-effects.
% abusive or threatening speech or writing that expresses prejudice against a particular group, especially on the basis of race, religion, or sexual orientation.
Hate speech, which is our focus in this paper, is usually understood as abusive or threatening speech or writing that expresses prejudice against particular groups.
It is a phenomenon related to human behaviour that spans across different cultures and languages, which can seriously limit the use of social platforms for groups like women, minorities and other vulnerable segments.
Furthermore, virtual-world hateful expressions can have the aggravated effect of sometimes translating into actual hate crimes in the physical world\footnote{https://time.com/5436809/twitter-apologizes-threat-mail-bomb-suspect/}\footnote{https://www.cbc.ca/news/canada/toronto/mosque-stabbing-suspect-1.5732078}. 
%

%%% NEED/PROBLEM (multilingual approaches) %%%
%
Automatic detection of hate speech messages in online social media platforms is an open and challenging multidimensional problem~\cite{arango2020hate}.
Despite the worldwide extent of this problem, limitations of existing solutions are even more profound when we consider that most of the research in this area has been centered in text written in English \cite{badjatiya2017deep, DBLP:conf/ecir/AgrawalA18, davidson2017automated}. 
This is also evidenced by the scarcity of models and learning resources (specific datasets, lexicons or word representations) for hate speech detection in languages other than English.
There have been some recent efforts towards systematically addressing the multilingual aspects of hate speech detection~\cite{DBLP:conf/semeval/Almatarneh0P19,DBLP:conf/semeval/BenitoAI19,Stappen2020,Saketh06465}.
Most of these works rely on emerging multilingual tools developed by the NLP community, in particular general-purpose multilingual text representations~\cite{Conneau2018,SchwenkD17,DevlinCLT19}. % agregar MUSE y las otroas cosas
These are representations for words or short texts that map input text data from several different languages into a single feature space.
However, as an emergent topic, there is still no consensus on how to satisfactorily undertake multilingual hate speech.
%in view of the lack of diverse NLP resources and labeled data for the specific problem.
%
Hence, it becomes imperative to find and compare approaches that are language-agnostic or that can take advantage of resources from more represented languages and apply them to languages with little to no resources.

We are particularly interested in \emph{cross-lingual} settings in which we have \emph{target language} for which it is assumed that there are no available resources (e.g., specific labeled data) for hate speech and one needs to leverage resources from a different language.
This setting is sometimes called \emph{zero-shot multilingual learning}, as one usually trains a model using data from one language, and then test it for a different target language without providing any labeled data for the target language.
Cross-lingual and language-independent approaches would facilitate transfer learning from English (and other languages) to low-resource languages.
%

%%% PROPOSED SOLUTION %%%
%
%We propose to address hate speech classification from a multilingual perspective by investigating cross-lingual and language independent representations.
%
We specifically analized in this paper which are the most effective alternatives for cross-lingual hate speech classification.
%, by both improving hate speech detection and by understanding cross-language hate patterns.
%
%In addition, we explore if models specifically designed for the multilingual hate speech domain improve classification, hence providing information about the context in which hateful terms occur.
%
We first consider approaches based on general-purpose multilingual text representations, in particular multilingual word embeddings~\cite{Conneau2018,SchwenkD17,DevlinCLT19}.
We hyphotesize that general-purpose multilingual word embeddings may not effectively capture some patterns that naturally arise when words are used in a hateful context, instead of a general context.
For instance, while some words related to nationality, religion, and race can be used in neutral contexts in general text, they can appear in very different contexts when used in hate speech, acquiring harmful meanings~\cite{ElSheriefKNWB18}. 

Motivated by the previous observation, we propose a set of {multilingual word embeddings} specifically created for hate speech. 
To achieve this we adopted the method proposed by \citeauthor{FaruquiD14} \cite{FaruquiD14}, which finds embedding projections that maximize the correlation between word embeddings from different feature spaces.
We use this method to align several monolingual hate speech embeddings independently created for each language in an unsupervised way.
%We emphasize that it is very important for the monolingual embeddings to be created by unsupervised training and not by relying of specific labeled data.
%
We evaluate the effectiveness of our approach in relation to other general-purpose representations by using them as input features for several machine learning models and on three different languages: English, Spanish and Italian. 

%
%%% FINDINGS AND IMPLICATIONS %%%
Our findings show that in general, the use of our hate specific representations improved cross-lingual models performance in comparison to general-purpose representations and pre-trained models. 
This suggests that besides the information provided by translating the general meaning of words to different languages, there are more specific cross-cutting patterns in how hate speech is displayed in those languages.
These patterns allows us to transfer knowledge from one language to another when detecting hate speech.
Along this line, we provide a qualitative analysis of our domain-specific multilingual embeddings by exploring word contexts in the three languages that we consider.
Our preliminary analysis show that hate specific embeddings are able to capture non-traditional translations of words from one language to other. 
For instance, if for a general purpose multilingual embedding the natural context-based translation (see Section \ref{sec:qualitative_evaluation} for details) of the Italian word ``{migranti}'' is ``migrants'' in English, and ``migrantes'' in Spanish, in our hate embedding the translations are ``illegals'' and ``palestinos'', respectively. 
We show several other classes of translations and discuss how they can be used not only for classification but also for better understanding of hate speech as a multicultural problem.

%% Object and CONTRIBUTIONS %%%
%
%In this paper, we present our proposed approach for cross-lingual hate speech detection, based on domain-specific word embeddings (i.e., hate speech embeddings), which allow us to leverage knowledge from different languages.
%
\textbf{ Contributions:}

1. We introduce the first domain-specific multilingual word representation (word embeddings) for hate speech classification.

2. We present a comprehensive evaluation of different approaches for cross-lingual hate speech detection.
 
3. We qualitatively show cross-lingual relations between terms in the context of hate speech derived from our domain-specific embeddings.

%

%%% Roadmap %%%

\noindent{\bf Warning:} Because of the topic that we consider, some example words mentioned in the paper may be considered offensive.

\noindent{\bf Reproducibility:} All of our code, experiments, and datasets, as well as our proposed word embeddings for hate speech will be publicly available in a centralized repository.

In the rest of the paper we first describe the related work on hate speech detection in monolingual and multilingual settings in Section~\ref{sec:related_work}. In Section~\ref{sec:methodology} we present our methodology including the standard models and input representations that we consider, introducing also our hate-specific word embeddings representations.
Our main quantitative and qualitative results are presented in Section~\ref{sec:specific_task_evaluation}.

\section{RELATED WORK}\label{sec:related_work}
In this section we review works related to hate speech detection 
in monolingual and multilingual scenarios, as well as methods for word embedding projections.
\subsection{Monolingual Hate Speech Detection}
Although the field of automatic hate speech detection have gained popularity in the past years, most of the existing approaches have been constructed for monolingual English scenarios. 
Several of them have approach the problem using \emph{traditional} machine-learning strategies~\cite{DBLP:conf/websci/ChatzakouKBCSV17,DBLP:conf/naacl/HeeaseemH16,davidson2017automated} and different types of representations mixing text representations with handcrafted features extracted from messages meta-information \cite{ papegnies2017graph, tahmasbi2018socio}. 
The English hate speech detection has also been addressed with deep-learning methods and word embedding representations~\cite{CNN1,CNN2,ESWC18,AISC18}. 
In addition, critical analysis of English systems and datasets have been conducted, helping to understand better the problem in English scenarios\cite{ArangoPP19, DavidsonBias, SapCGCS19}.

Other languages such as Spanish \cite{Pereira-Kohatsu19}, Italian \cite{SanguinettiEtAlLREC2018}, and Portuguese \cite{fortuna2019}, have been addressed using similar techniques as in the English approximations, though in a fewer amount of works which might be in part due to the lack of available resources.

\subsection{Cross-lingual Hate Speech Detection.}
As it has been shown for other tasks~\cite{Conneau2018, ConneauRLWBSS18}, 
a multilingual approximation to the hate speech problem would help to advance the state of the art for under-represented languages.
For languages with little to no results, one would need approaches in which 
no information about the target language (the one over which one wants to detect hate) is used during the training process.
We refer to this constrained scenario as \emph{cross-lingual}.
%approach as a particular type of multilingual approach where none information about the target language is used during the training process.
%
In spite of several approaches designed for dealing with hate speech in different languages~\cite{DBLP:conf/semeval/Almatarneh0P19,DBLP:conf/semeval/BenitoAI19,DBLP:conf/semeval/RaiyaniGQN19}, there are only a few reports on strict cross-lingual evaluation on the recent related literature~\cite{lan2020focused,Stappen2020,Saketh06465}.

Translating all the data to a common language as strategy is one of the strategies that have been applied. 
\citeauthor{PamungkasP19} \cite{PamungkasP19}, explored this strategy using English, Spanish, Italian and Deutsch datasets. Once all data is in the same language, a monolingual strategy is applied. 
%Another approach is to train the models with the training data and its translation using MUSE \footnote{https://github.com/facebookresearch/MUSE} multilingual embeddings.
%These representations are used along with LSTM based architectures. The best result (68.60\%) was obtained in a English\-Spanish evaluation using MT, Muse embeddings and a LSTM architecture.

Meta-information from the network dynamics of the message and the message authors could be considered ``multilingual'' features as they are not directly related to the language in which the text is written. 
This type of features is used by~\citet{arango2020hate} where they are combined with traditional machine-learning models in cross-lingual evaluation for English and Spanish datasets.

\citeauthor{Saketh06465} \cite{Saketh06465} experimented in a cross-lingual (zero-shot and few-shot learning) manner over nine different languages. They tested different combinations of different machine learning models and vector representations such as MUSE\footnote{https://github.com/facebookresearch/MUSE} and LASER\footnote{https://github.com/facebookresearch/LASER} embeddings. 
The combination of LASER and a Logistic Regression (LR) model turned out to be the best combination in most of the experimental setups. 
This show that traditional machine learning models still have to be considered for this task.
%In spite of several  DNN architectures were tested, one of the best result (68.61\%) was obtained using a Logistic Regression models with LASER embeddings. That is an evidence that the type of models that better solve the problem is still not clear.

Extracting features from pre-trained models is the strategy followed by \citeauthor{Stappen2020} \cite{Stappen2020}.
The authors proposed an architecture where initially the tokenized text is propagated through a pre-trained model, extracting vector representations. 
These representations are fed into a classification model. 
%The best result (65.04\%) was obtained in a English-Spanish evaluation
Fine-tuning pre-trained multilingual models like BERT\footnote{https://github.com/huggingface/transformers} over the training data as an end-to-end classification model is another strategy that have been used also in the related literature~\cite{Stappen2020,ANDES}.

The best performance reported by each of them was achieved using different representations and different models. 
There are not categorical conclusions about which model perform better for this scenario. 
The cross-lingual evaluation of hate speech is still a new research area, and the results are just a few. 
The different approximations usually rely on general-purpose representations and models.
We did not find works reporting efforts to construct specialized representations (word embeddings) for this problem as is one of the focus of our paper.

\subsection{Monolingual Specific Domain Word Embeddings.}
In the related literature can be found some works describing the construction of specific-domain word embeddings for hate speech detection, but only in monolingual scenarios. 
As far as we know, there are not reported efforts for constructing multi-lingual specific domain word embeddings.

\citet{KambleHindiWE} constructed word embeddings using a Word2Vec model Hindi-English code-mixed tweets from the hate speech domain. 
To show their effectiveness they trained different deep-learning hate speech classifiers. \citet{AlatawiSupremacist} also describe the construction of Word2Vec word-embeddings in the domain of English white supremacist hate speech. 
The authors perform qualitative and quantitative comparison of these embeddings qith other from general domain. In both cases the specific-domain corpus is obtained using previously kwon hateful terms as queries.
\citet{badjatiya2017deep} propose the construction of English word embeddings for the specific domain of hate speech using a labeled dataset and an LSTM-based model. However, their validation strategy was considered wrong since the same dataset was used for constructing embeddings and validation purposes \cite{ArangoPP19}.

\subsection{Projection-Based Multilingual\\ Word Embeddings}
Although some specific techniques can be applied for constructing multilingual embedding for specific-tasks, the \emph{projection technique} represents an attractive option as it requires resources relatively easy to obtain for most of the tasks \cite{GlavasLRV19}.
The general idea is to linearly project two vector spaces into a common one by optimizing the relationship between dictionary-paired vectors obtained from bilingual dictionaries.
The bilingual dictionaries can be induced from the data (unsupervised methods) \cite{Conneau2018,LampleCRDJ18, ArtetxeLA18} or provided beforehand (supervised methods) \cite{MikolovLS13, FaruquiD14,joulin2018loss,ruder2018}. 
Different projection techniques have been proposed in the related literature defining different optimization problems such as minimizing distances between equivalent vectors  \cite{MikolovLS13}, minimizing cosine distance modifications \cite{joulin2018loss}, maximizing correlations between equivalent terms \cite{FaruquiD14} among others. 
\citet{RuderVS19} presented a complete survey about different types of alignments.  
As far as we know, none of these techniques have been applied for creating multilingual embeddings for the hate speech domain.
\section{Methodology}\label{sec:methodology}
 With the goal of analyzing which are the most effective alternatives for multilingual hate speech classification, we evaluate different settings by combining diverse input representations (language independent features, word and sentence embeddings) and several models (traditional machine learning and deep learning models) that use those representations to solve a cross-lingual task.
These experiments allow us to evaluate how our hate speech word representations (that we will introduce in Section \ref{sec:HateEmb}) perform in comparison with the general-purpose ones. 
%
%For understanding cross-language hateful patterns, we analyze relations among terms that could help us to understand better the phenomenon of hate, and evaluate in a qualitative manner our hate speech representations.
%
As we are interested in the cross-lingual scenario, we conducted our experiments over three datasets in different languages: English, Spanish and Italian.

\subsection{Datasets} \label{sec:datasets}
\if False
Despite of the fact there are more resources available for the English language, it was difficult to find datasets with all the available information we wanted to use even in English.
Most of the publicly available hate speech datasets only contain the text of the message with no other meta-information.
This issue reduces our possibilities to exploit meta-information usually included in social media data, such as user-specific information, information about the network of the user, and context information about specific messages. 
It has been shown in the literature that at least some anonymized data about user is needed to avoid ``user bias''~\cite{ArangoPP19}.
Considering these requirements, we selected Twitter datasets that contain all the needed information from three different languages: English, Spanish and Italian. A summary of the datasets statistics can  be found in Table  \ref{tab: datasets}.
\fi

For the purpose of validating our results we needed annotated datasets.
The availability of annotated datasets (specially for non-English languages) is very poor and we are aware of this limitation. With our proposal we intent to make the best of very poor resources.

For the English language we mixed two different datasets. The first one was constructed by \citet{ArangoPP19}.
For constructing this dataset, the authors combined two previously published datasets \cite{waseem2016you, davidson2017automated}.
%with the objective of reducing a previously existing user bias. 
%
The types of hateful content addressed in this dataset are racism, sexism and xenophobia.
We consider these three classes as hateful, thus having a final dataset with only two labels: hateful, with 2,920 texts, and non hateful with 5,576.

The tweets on this dataset were originated in the context of hate in the United States of America. 
Therefore the targets of hate, as well as specific terms, are framed on that particular cultural context.

We also considered the English Twitter dataset proposed by \citet{SemEval19}  for hate speech against immigrants and women, therefore the targets of hate are similar to the one in the \citet{ArangoPP19} dataset. Each tweet is tagged as either hate (5,390) speech or normal (7,415).

For the Italian language we consider the dataset described by \citet{SanguinettiEtAlLREC2018}, which is part of a hate speech Italian monitoring program. 
%
%The authors made the tweet-IDs public so it was possible for us to obtain all the meta-information of the tweets (at least for those tweets which were not deleted from Twitter). 
%
This dataset includes 1,291 tweets expressing hate against immigrants and other 5,637 negative examples.

The Spanish dataset constructed by \citet{Pereira-Kohatsu19} we  is composed of tweets addressing topics of racism, sexism and xenophobia. 
We use it as a binary dataset where 1,576 of the tweets are labeled as hateful and 4,434  as non hateful. 
The authors of the dataset recovered tweets specifically originated in Spain.
In order of collecting more labeled examples we also considered the Spanish Twitter dataset proposed by \citet{SemEval19} composed by 2,228 addressing hate speech against inmigrant and women; and 3,137 non-hateful tweets.

A summary of the datasets statistics can  be found in Table  \ref{tab: datasets}.
\input{Datasets}

\subsection{General-Purpose Multilingual Representations}\label{sec:data_representations}
Recall that our input is composed of short texts from social media (tweets). The metadata information present in those tweets could be considered as features, but unfortunately is only available in very few of the available datasets. Therefore we considered only text representations that we describe next.

\if True
\subsubsection{Language-Independent Features (LIF)}\label{sec:Lif}
We extracted seventeen handcrafted features that have been shown useful in monolingual hate speech detection \cite{DBLP:conf/websci/ChatzakouKBCSV17,DBLP:conf/mm/Huang0A14, papegnies2017graph, tahmasbi2018socio}.
Some of these features have also proved useful in cross-lingual settings~\cite{arango2020hate}.
The motivation for using this type of features in our context is that they are not directly related to a specific language.

Some of these features were extracted from the content of the tweets. We counted the occurrences of different elements from the text: hashtags, user mentions, exclamation marks,  multimedia objects in the tweet, capital letters, urls, sad and happy emoticons\footnote{http://kt.ijs.si/data/Emoji\_sentiment\_ranking/} and also the length of the tweets.
In addition we used features related to the structure of the network the tweets appeared on. From each tweet we counted the number of retweets, responses and likes. Using tweet author meta-information we also quantified the number of followers and friends, lists the user is subscribed to, tweets the user has liked and tweets the user has written.
In what follows we call LIF to this representation of the input data.

\subsubsection{General-Purpose Multilingual Word Embeddings}\label{sec:gpembeddings}
\fi

We consider three types of multilingual embeddings: MUSE~\cite{Conneau2018},
BERT~\cite{DevlinCLT19} and LASER~\cite{SchwenkD17}.
MUSE is a set of general embeddings aligned for multilingual contexts.
BERT is a general purpose pre-trained model for NLP that can be used to produce embeddings for sentences (sequences of words).
BERT can be trained in an unsupervised way from big corpora, and the authors of the original model provided a BERT version trained over a corpus containing text from 104 languages~\cite{DevlinCLT19}. 
From now on we call it multilingual BERT (or mBERT for short).
In some of the monolingual results, we also consider the monolingual versions of BERT and fine tune it for the specific task (we refer the reader to~\cite{DevlinCLT19} for details on fine tuning BERT).
LASER~\cite{SchwenkD17} is a recently proposed model to produce multilingual embeddings for sentences.
As opposed to mBERT, LASER was constructed specifically for the multilingual context.
Although the three models has recently been used on multilingual hate speech detection \cite{BojkovskyP19, Saketh06465},
there is still no consensus about which of these representations perform better for this task. 
%
%In general, the reported results are lower in comparison with the ones reached in mono-lingual scenarios. 
%
We evaluate the usefulness of these three representations using them to construct input features for several classification models. 

\subsection{Hate-Speech-Specific Word Embeddings}\label{sec:HateEmb}
Besides the standard representations described in the previous sections and since the phenomenon of hate speech has its own characteristics, we also consider a domain-specific representation. 
%To prove the importance of this, 
Thus, we constructed specific word representations using a projection embedding technique with the following general steps: \textit{(1)} constructing monolingual vector spaces for each language separately in an unsupervised way, \textit{(2)} preparing a bilingual dictionary for each pair of languages, and \textit{(3)} aligning the monolingual spaces into a single embedding space.
This method has the advantages of being independent of the algorithm used for constructing the monolingual embeddings, and only requiring a bilingual dictionary instead of a big amount of parallel or labelled data~\cite{RuderVS19,ArtetxeRY20}.

We next describe every one of the above mentioned steps in more detail.

\subsubsection{Monolingual vector spaces}
Using the Twitter API\footnote{https://github.com/tweepy/tweepy}, we recovered tweets for every language by simply using some general hateful terms as queries.
The seeds would guarantee the existence of hateful terms in the resulting word embedding vocabulary and a higher probability for hateful tweets to appear in the corpora compared with recovering them randomly.
The sizes of the corpora are: ~30M English,~10M Spanish and Italian each.
%The goal in this step is constructing embeddings that capture better the context of the words in hateful scenarios. 
%
     
In our specific context, the hateful terms were obtained from the multilingual Hurtlex lexicon~\cite{BassignanaBP18}\footnote{https://github.com/valeriobasile/hurtlex} and an English lexicon constructed by~ \citet{davidson2017automated}\footnote{https://github.com/t-davidson/hate-speech-and-offensive-language/blob/master/lexicons/refined\_ngram\_dict.csv}.
With the recovered data, we trained Word2vec models  for each individual language (English, Spanish, and Italian).

We emphasize that the process described here is a simple process that can essentially be replicated for any language by only having a set of hateful words, without the need of any data specifically labelled for hate speech.
This is a necessary condition to be applicable in low resource contexts.

\subsubsection{Bilingual dictionary}
As bilingual dictionary we used word-aligned pairs from Hurtlex~ \cite{BassignanaBP18}. 
The Hurtlex multilingual lexicon helped us to match hateful terms between different languages. 
Acording to \citeauthor{Shakurova} \cite{Shakurova} better results are obtained when the bilingual lexicon is from the specific domain of the task.
The particularity of Hurtlex is that it includes terms that have different colloquial equivalences that are not usually included in generic dictionaries, as well as words that could appear usually in hateful content.
As an example we show in Table \ref{tab:bilingual_lexicon} the bilingual equivalences of the English word ``pussy''. 

\input{bilingual_lexicon}
 
\subsubsection{Aligning the monolingual spaces}
As alignment algorithm, we  adopted  the framework proposed by  \citet{FaruquiD14}\footnote{https://github.com/mfaruqui/crosslingual\-cca} since it have been applied to different domain-specific tasks such as: sequence labelling for 
Curriculum Vitae parsing \cite{Shakurova}, text categorization \cite{AmmarMTLDS16}, cross-lingual information retrieval, and document classification \cite{GlavasLRV19}.
 In this process, a pair of monolingual word vectors are projected into a common space by learning two projection matrices $V$ and $W$ that maximize the correlation between the dictionary-paired projected vectors.
 
 \begin{figure}[!h]
   \centering
   \begin{tabular}{@{}c@{\hspace{.5cm}}c@{}}
       \includegraphics[page=1,width=.45\textwidth]{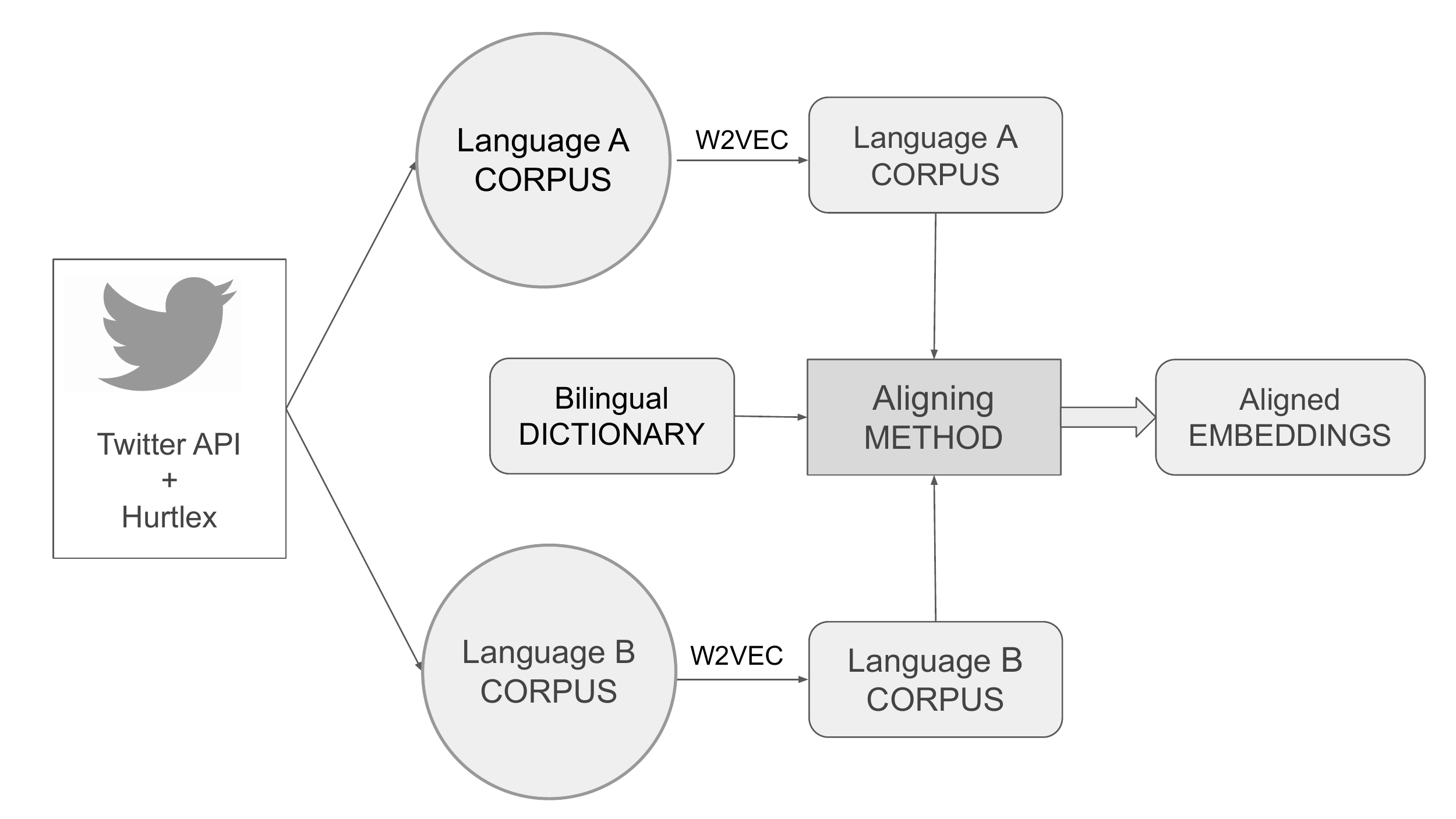} \\
   \end{tabular}
 \caption{Constructing process of the Hate-Speech-Specific Word Embeddings.}
 \label{fig:Test}
\end{figure}
% \includegraphics[width=8cm, height=4.5cm]{images/aligning proces.jpg}

%%Let $X_{L1}^{'}\subseteq X_{L1}$ where every word to other in $X_{L2}^{'}\subseteq X_{L2}$ 
%\begin{equation}
%V,W = CCA(X_{L1}^{'},X_{L2}^{'})
%\end{equation}
%\begin{equation}
%X_{L1}^{*} = X_{L1}V, \  X_{L2}^{*} = X_{L2}W
%\end{equation}
%\begin{equation}
%= {max}_{\tiny{V,W}}\ \rho(X_{L1}^{'}V,X_{L2}^{'}W)
%\end{equation}

\subsection{Models for Hate Speech Classification}\label{sec:clasificarion_models}
In the current research on the cross-lingual detection subject, different methods have been used with similar results. Since the best model for approaching the problem is still not clear, we perform experiments with several methods including traditional machine learning methods and deep learning models. 
As traditional machine learning models we used Logistic Regression, XGBoost (XGB), Support Vector Machines, Random Forest (RF), Decision Trees (DT), and Naive Bayes classifiers . 

Regarding the deep learning models that we considered, we tested: Convolutional Neural Networks (CNN), Feedforward Neural Networks (FNN), Long Short-Term Memory networks (LSTM). 
In addition, we combined LSTM and CNN layers (LSTMCNN), as well as LSTM with Attention \cite{attention} (LSTMATTN). 
All these models were tuned in order to find the best possible values for the different hyperparameter combinations. 
We also performed fine-tuning of the corresponding monolingual BERT models for Italian, \footnote{https://github.com/marcopoli/AlBERTo-it}, Spanish\footnote{https://github.com/dccuchile/beto} and English\footnote{https://github.com/huggingface/transformers} for conducted some monolingual experiments as reference for the cross-lingual ones.
\section{Evaluation and Results}\label{sec:specific_task_evaluation}
We tested several combinations of models and input representations 
for hate speech detection.
%
%These experiments would lead us to find which strategy perform better for hate speech detection, as well as testing the usefulness of our specific task word embedding (HateEmb) for hate-speech detection.
%
We evaluated the different combinations in both, monolingual (Secction \ref{sec:results_monolingual}) and cross-lingual (Secction \ref{sec:results_crosslingual}) scenarios. 
The available data in the three different languagesdatasets (Section \ref{sec:datasets}) were portioned into training, development and testing sets. 
These partitions remained the same for all the experiments.

The traditional machine learning models were combined with  LASER sentence embeddingss. 
On the other hand, DNN architectures were combined with word embeddings extracted from pre-trained monolingual and multilingual BERT (mBERT), MUSE multilingual word embeddings and our hateful multilingual word embeddings (HateEmb).
\subsection{Monolingual Evaluation}\label{sec:results_monolingual}

\input{monolingual_summary}

Despite the fact that our main goal is the cross-lingual evaluation of hate speech detection, we considered important to perform comparisons of cross-lingual results with their monolingual counterparts. 
Intuitively, the closer the cross-lingual results are to the monolingual ones, the better they were able to transfer knowledge satisfactorily from one language to another.  

In Table \ref{tab:monolingual_summary}, we show the results (in terms of F-socre) obtained in a monolingual evaluation using different word (and sentence) embeddings representations combined with different DNN models.
At each column header we show how input features are computed (BERT, MUSE, HateEmb, etc.).
An exception is the ``Fine-tuned BERT'' column in which the task was solved end to end by adapting the parameters of a pre-trained BERT model to the hate speech specific task.
We only report the best result over all models that we tested, and over each result we depict the model used to obtain that result (for instance, the best result for the English language using BERT input features was obtained by using an LSTM network with Attention).

In this monolingual case we can observe that, as expected, using multilingual input representations (mBERT, MUSE and HateEmb) did not improve classification results.
Moreover, for the three datasets that we consider, the BERT-based models show the best performances.
%
%
%The difference is more noticeable in Spanish and Italian languages where the use of monolingual BERT embeddings and the fine tuned model showed better results.
Another important observation is that the use of our hate-speech-specific word embeddings (HateEmb), although did not present the best results, they show results similar to other multilingual input representations.
Actually, if we only compare the results of the multilingual input representations, HateEmb surpass MUSE in the three languages and also mBERT for the Italian language.

\subsection{Cross-Lingual Evaluation}\label{sec:results_crosslingual}

\input{Cross_lingual_summary}

%In Table \ref{tab:Cross_lingual-DNN}, we show only the best results per representation for each of the experimental configurations.
%

Table \ref{tab:Cross_lingual-DNN} shows the results of cross-lingual experiments using several different input representations. 
In these experiments we first picked an input representation (shown in column headers as mBERT, MUSE, LASER and HateEmb in Table~\ref{tab:Cross_lingual-DNN}), then train a classifier using a source language (second column), and finally test it over a different target language (first column).
As we have mentioned, this setting is sometimes referred as \emph{zero-shot multilingual transfer learning}, as no data of the target language is presented during training.
This is, arguably, the most challenging multilingual transfer task.
We tested several different models over the input representations, and we report the one that gave the best result for every combination.

As we expected, the results in the cross-lingual setting are lower than the ones obtained in the monolingual evaluation.
The use of our proposed HateEmb embeddings show promising results, obtaining the best results in four of the six configurations with a considerable margin in several of them. 
We show the difference between the results of our embeddings compared with the best alternative method in the last column of Table~\ref{tab:Cross_lingual-DNN}.
Our HateEmb is surpassed only by configurations in which the multitlingual BERT model is used to obtain input features.
In that cases our proposal is the second best.
It should be noticed that BERT is a huge model with millions of parameters and needing specialized hardware to be trained.
In contrast, our embeddings are really lightweight and can be trained in general purpose machines.

The models that worked better in combination with the HateEmb embeddings are the ones based on LSTM architectures.
%
% The LIF input representations showed discrete results, but in some cases outperforms the use of word embeddings (e.g. Italian as source language and Spanish as target).
%
From all the traditional machine learning models used, the best performances were obtained by using an XGBoost classifiers for the cases of LASER input representations.
The best DNN model varies depending on the experiment, therefore conclusions about the best performing DNN for this task can not be taken. 

Several hyper-parameters were tested in an exhaustive hyperparameter tuning process. The best hyper-parameters were different depending on the cross-lingual setup and model. Since they are many, for the sake of space, we describe all in the code repository (to be publicly available).

%The overall best result (69.87\% F-score) was obtained when using the Spanish dataset as the training set with HateEmb and an LSTM architecture over the English set as testing.

The fact that our hateful embeddings were competitive with the more sophisticated but general ones, lead us to the hypothesis that we were able to capture important semantic information about hate-speech when training and aligning our embeddings for different languages.
We qualitatively asses a related hypothesis in the following section.

\subsection{Qualitative Evaluation of Hate Embeddings}\label{sec:qualitative_evaluation}

% Decir que mostramos la efectividad de los HateEmb

The intrinsic quality of multilingual word embeddings is usually evaluated on the Bilingual Lexicon Induction (BLI) task \cite{VulicGRK19, RuderVS19}. 
This task measures how well the vectors representing translations in different languages are close to each other
in the common embedding space. 
BLI relies on nearest neighbor search in the multilingual embedding space identifying the most similar word in the target language given a word in a source language.
The target and source words are expected to be translations for each other according to a validation dictionary~\cite{GlavasLRV19}.

We have several difficulties for using a BLI-like quantitative method to asses the intrinsic quality of our embeddings. 
The main difficulty is that, as we have argued before, we consider that hate speech is a problem where word meanings could go well beyond literal translations. 
Thus, having a low BLI score for general terms would not necessarily mean a low quality for hate speech detection.
Moreover, we already used the translation of some specific hate speech terms as bilingual dictionary when constructing our embeddings (see Section \ref{sec:HateEmb}).
Thus, using that same bilingual dictionary as a validation set for our embeddings would be meaningless.
One possible option would be to manually construct new hate speech specific bilingual dictionaries for evaluation which would be a highly time-consuming process. 
Moreover, being the hate speech problem a culture phenomenon, to go beyond standard trivial translations of words, one would need experts in the cultural use of complicated terms in different languages. 
 
Therefore, we decided not to measure the quality of our embedding based on quantitative tasks, but instead we present more specific qualitative analysis along the same idea of BLI-like tests.
The analysis gives some evidence that our created embeddings have the potential to map terms that are similarly used for hate speech in different languages.

%There is certain evidence that there is not always a direct relation between the performance of the embeddings in the BLI-task and the performance in downstream tasks~\cite{GlavasLRV19}.
%
%In addition, as we have argued before, we consider that hate speech is a problem where word meanings could go well beyond literal translations. 
%Our evaluation consists in observing the most related terms across-languages hoping to find equivalent hateful meanings and not necessarily their direct translation in both vector spaces and within the labeled datasets.
%
%While we evaluate the embedding, this action could also help us to understand the problem of hate speech better; How the phenomenon is expressed in different languages and therefore cultures.

\subsubsection{Cross-lingual relations in vectors spaces}\label{qualitative_vectors}
\input{neighbors_in_vectors}

Our first qualitative evaluation is inspired by BLI and consists in, giving a seed term, observing the most related terms across-languages.
%hoping to find equivalent hateful meanings and not necessarily their direct translation in both vector spaces and within the labeled datasets.
In Table \ref{tab:neighbors_in_vectors}, we show a sample of the relations between the terms comparing our embeddings (HateEmb) with general purpose multilingual embeddings (MUSE).
These terms were manually selected trying to represent some groups that might be the focus of hate, and ensuring their equivalences were not present in the bilingual dictionary that we use for aligning our embeddings (thus showing a \emph{new} relation).

For each selected \textit{source term}, we show the nearest neighbor (NN) embedding in the common space corresponding to a word in a language different to the source language.
In most cases the nearest neighbors in the MUSE space are terms which standard meanings are the same in both languages.
For example, for the Italian word ``migranti'', we found that the nearest terms on the MUSE space is ``migrants'' in English and ``migrantes'' in Spanish. 
On the other hand, the nearest neighbors on the HateEmb space are ``illegals'' and ``palestinos'' words in English and Spanish, respectively, whose standard (neutral) translations do not match with the source word. 
Although not direct translations of each other, we argue that these words are likely to appear in similar context in hateful scenarios in the languages that we considered. 

We consider that evaluating the hate speech specific embeddings considering literal translations such as "migranti" and "migrante"  it is not suitable, since in the hateful content the word "migranti" could adquire different meanings.
The right equivalence to expect is not known, due to the complexity of the hate speech problem.
Moreover, expecting same relations in different languages (e.g. "migrants" - "terrorist" = "migrantes" - "terroristas") would be also wrong. 
The targets of hate in different languages are different depending on the socio-cultural scenario.
We prefer to extract information from vector spaces and datasets, and qualitatively evaluate the observed relations. 

In most of the cases we were able to observed non-trivial translations when exploring our hateful embeddings, though in some others we could observe that the equivalences are the same as in MUSE (e.g. ``negros'' in Spanish, as ``blacks'' and ``neri'' in English and Italian, respectively). 

%
%From the differences between equivalent terms in general-purpose and hateful vector spaces, we conclude that hate speech, as other specific domain has its own characteristics: neutrally connoted words can acquire hateful meanings.
%
%This characteristic makes a hate speech detection a complex task, and as better those cross-lingual meanings are represented, as better cross-lingual classification can be conducted.

More experimentation is definitely needed to derive a more robust conclusion, but we think that the qualitative results presented here are a positive evidence on how our domain-specific embeddings are capturing non-trivial meanings and translations.

\subsubsection{Cross-lingual relations in labeled datasets}

\input{neighbors_in_data}

In this section we qualitatively explore the ability of our embeddings to capture equivalences between hateful concepts in different languages over a labeled dataset.
In the previous section we use similarity measures (nearest neighbors) over the general embedding space and for all the vocabulary used to construct those embeddings (unlabeled data).
In contrast, in this section we focus on texts from the positive class of the hate speech labeled datasets in different languages.
That is, we focus on multilingual data that we know that contains hateful information.
We use our domain-specific hate embeddings plus \textit{association rules} to devise a similarity measure among terms in different languages as a way to obtaining new and more specific translations for complicated hate concepts. 
The motivation for this experiment is twofold. 
On the one hand this would serve as an intrinsic qualitative evaluation as we can asses how good are the translations obtained for simple hateful terms.
On the other hand we expect that this experiment allows us to preliminary obtain a more rich set of equivalences regarding hate in different languages.

We next explain in more detail the method we devised to obtain the equivalences.
For the first step, let $x$ be a word and $U$ a set of words all from one of the labeled datasets. 
From each dataset, we extract association rules of the form $\{x\}\Rightarrow U$ with the following semantics: if $x$ occurs in a text $T$ (tweet in our case), then $U\subseteq T$ with certain confidence \cite{HippGN00}.
In that way we can find words that usually occur together in the same text.
We extracted rules for the top most frequent terms $x$ in each dataset and we measure the strength of the rules using the standard \textit{support} and \textit{confidence} metrics.
All the rules extracted were refined by imposing lower bounds in confidence and support \cite{jain2012hiding}.

We note that in each dataset and for each frequent term $x$, one can obtain many different association rules.
Using all the rules with the form $\{x\} \Rightarrow U_i$, we computed \emph{the context} of the word $x$ as $C(x) = \bigcup U_i$.
Finally, our similarity measure for two words is based on the similarity of contexts for those words.
We still need to introduce an additional notation before presenting our similarity measure.
For every word $u\in C(x)$ we denote by $\textit{supp}_x(u)$ and $\textit{conf}_x(u)$ the support and confidence of term $u$ in the association rule of the form $\{x\}\Rightarrow U$ it appears. 
Given words $u$ and $v$, appearing in 
contexts, say $C(x)$ and $C(y)$, respectively, we define the following expression that essentially compares their support and confidence metrics in their respective contexts:
\[1-|\textit{supp}_x(u)-\textit{supp}_y(v)|/2 + |\textit{conf}_x(u)-\textit{conf}_y(v)|/2\]
We denote this expression simply by $\textit{met-sim}(u,v)$.
Finally, we can combine the above similarity for context words with an usual embedding similarity based on cosine similarity by averaging both:
\[
\textit{sim}(u,v) = (\textit{cos-sim}(u,v) + \textit{met-sim}(u,v))/2
\]
That is, we give the same importance to how the vectors are similar across the multilingual vector spaces (cosine similarity), but also how they have similarly importance in the association rules they appear in.

We have all the necessary ingredients to define the \emph{context similarity} of words. Let $x$ and $y$ be words (possibly from datasets in different languages) with contexts $A=C(x)$ and $B=C(y)$.
Then their context similarity, denoted by $\textit{cont-sim}(x,y)$ is defined as
\[
\frac{1}{2}\bigg(\mean_{u\in A}\Big(\max_{v\in B} \textit{sim}(u,v)\Big) +
\mean_{v\in B}\Big(\max_{u\in A} \textit{sim}(u,v)\Big)\bigg)
\]
That is, for every word in $x$'s context ($A$) we compute its maximum similarity with words in $y$'s context ($B$) and take the mean over all those similarities, and the other way around (mean over $B$ of the maxium similarities with words in $A$), and the results of both directions are averaged. %\ref{tab:Cualitative_evaluation}.
%%% tiene sentido agregar alguna intuición?

%
%For each top frequent word $u$, we defined its context $A$ in the dataset as the set of words that frequently appear with it in the same tweets.
%
%Using all the rules with form $\{u\} \Rightarrow U_i$, we computed the context of the word $u$ as $A = \bigcup U_i$.
%

%We designed a similarity measure that is focused on computing the similarity between two context rather than between two words.
%%
%Being $A$ and $B$ the contexts of two words $u$ and $v$ their similarity is calculated as:
%\begin{equation}\label{eq1}
%Sim\left (A,B\right) = (Mean_{x\in A}sim(x,B) + Mean_{y\in B} sim(y,A))/ 2
%\end{equation}

%In that way we calculate the mean of the similarities among of all the words in $A$ with the words in $B$ (first addend in (\ref{eq1})) and the other way around (second addend).
%%
%The results of both directions are averaged. \ref{tab:Cualitative_evaluation}.
%\begin{equation}\label{eq2}
%sim(x,S) = Max_{y\in S}((s_{cos}(x,y) + s_{met}(x,y))/2)
%\end{equation}
%We defined the similarity between one word ($x$) and one set of words ($S$) taking into account similarity between vectors with cosine similarity ($s_{cos}$) and similarities between the metrics \textit{confidence} and \textit{support} ($s_{met}$). 
%%
%That is, we give the same importance to how their vectors are similar across the multilingual vector spaces, but also how they have similar importance in the rules they appear. 
%\begin{equation}
%s_{met}(x,y) = 1 - (|sup(x) - sup(y)| + |conf(x) - conf(y)|)/2
%\end{equation}

We use our embeddings and the above defined context-based similarity to perform the following experiment over the labeled datasets.
For each dataset of every language, we first selected some frequent words appearing in the hateful-labeled texts.
We call them \emph{seed terms}.
Then for each seed term we selected the words that are more (context-) similar over all the words appearing in hateful labeled texts in a different language.
Table~\ref{tab:Cualitative_evaluation} shows examples of seed terms and the top three most similar words (with their respective similarity score).
As a comparison the table also shows the experiment for the same seed terms but considering the most similar word over the non-hate texts.

%While in Section \ref{qualitative_vectors} we analyzed relations between vector spaces, in this evaluation we considered relations across datasets, which could express more specific information about the type of hate speech treated and the particularities of the culture from the countries the datasets were extracted. 
 \input{neighbors_in_data_MUSE}

Despite of the fact that the labeled datasets are relatively small and from specific types of hate, we still find interesting cross-lingual relations. 
As expected these relations are different depending on the nature of text, that is hateful or non-hateful. 
For example, for the Spanish word ``terroristi'', we found the words ``muslims'' and ``fascistas'' as the most similar words in the hateful texts in English and Spanish, respectively.

That means that, according to our similarity function, the word ``muslims'' appears in similar contexts in English as "terroristi" in Italian. (e.g. muslims race, idiot, cult murder terrorism.) 

These relations can be interpreted as a cross-cultural similarity in the way this two concepts are, in a similar way, part of the hate speech phenomenon. 

%
% On the other hand in the non-hateful class we found as most similar for ``gitano'' two neutrally connoted words ``channel'' and ``viene''.
% %
We emphasize that these relationships although only qualitative, cannot be so clearly found when one perform the same experiment using other general purpose multilingual embeddings (see Table~\ref{tab:Cualitative_evaluation_MUSE} in the Appendix).
%(see Table~\ref{tab:Cualitative_evaluation_MUSE} in the Appendix).

%\subsection{Discussion}\label{sec:discussion}
\section{Discussion and Concluding Remarks}\label{sec:conclusions and future work}
%The experiments we conducted for a specific-task evaluation of our hate speech embeddings (HateEmb) were separated in monolingual and cross-lingual scenarios.
%In a monolingual evaluation, the results did not outperform the ones obtained using BERT embeddings, though they were still competitive. 
%
%Therefore their use in monolingual scenarios could be an interesting alternative when lower dimensionality is required.

We have presented a detailed analysis of cross-lingual hate speech approaches, with the goal of transferring knowledge from one (or more) languages to another.

Our results indicate that there indeed are cross-cutting patterns in hate speech that span different languages.
In particular, in our cross-lingual evaluation setup our domain specific HateEmb embeddings show the best overall results.
%, though in was not the best representation alternative for all configurations. 
%

Their competitiveness with other more sophisticated general-purpose representations is a sign that they are able to capture important specific domain information.

%
%The LIF we used, were not enough for transferring knowledge from one language to another. 
%
%However for some configurations their use combined with traditional machine learning performed better than the use of MUSE and DNNs. 
%
%We concluded that other LIF have to be explored in order to improve the results with this approach.

%
We also performed a qualitative exploratory analysis, which showed cross-lingual relations in our vector spaces and within datasets, supporting the observation that the context of words in a hateful scenario is very different from the context of the same words in a general scenario.
This validates the importance of specific-domain representations for the hate-speech detection problem. 
As our results show, the construction of domain-specific hate-speech word embeddings can be a key tool to further explore in cross-lingual scenarios. 
We expect that transferring knowledge from one language to another in hate speech detection, will contribute to the development of better models for this task from a multilingual perspective.
Thus, improving the diversity of application scenarios to other languages spoken worldwide without requiring massive amounts of labeled data. 

For future work, we are interested in exploring other algorithms for constructing specific representations for this task, since it seems a promising way to improve classification results.

% The similarity function we used to qualitative our hate specific embeddings, could be incorporated in the classification process as it seems to capture similarities across languages.

%\section{Con}
%\input{AugmentedData}
\balance

\appendix
\section{APPENDIX}

\subsection{Similarities across datasets using MUSE embeddings.}\label{sec:appendix1}
Table~\ref{tab:Cualitative_evaluation_MUSE} shows results for an experiment similar to the one presented in Table~\ref{tab:Cualitative_evaluation} but considering the general purpose MUSE multilingual embeddings instead of our hate-specific embeddings.

% \subsection{Hyperparameters}
% After an hyperparameter tuning process we found the hyperparameters that work better for each model in the cross-lingual settings. In Table \ref{hiper} we show the hyperparameters of the models that result to be better on each experimental setup.
% \input{hyper}
%\bibliographystyle{unsrt}
%\bibliographystyle{plainnat}
\bibliography{bibliography}
%\printbibliography

\end{document}

%% file: Datasets.tex
% \usepackage[normalem]{ulem}
\begin{table}\caption{Description of the datasets we used for hate speech evaluation. For each dataset  we show the number of tweets per class.}\label{tab: datasets}
\begin{tabular}{lcccc}
     \small{Language} & \small{Dataset} & \small{Hate} & \small{Non-Hate} & \small{Total} \\
     
\midrule

English    &\small{Arango et al. \cite{ArangoPP19}}  & 1,490 &  5,576 &  7,006  \\ 
& Basile et al. \cite{SemEval19} & 5,390  &7,415  & 12,805   \\ 
\midrule
& Basile et al. \cite{SemEval19}  &2,228 &3,137& 5,365   \\ 
Spanish   & Pereira et al. \cite{Pereira-Kohatsu19} & 1,576  &4,434 &  6,000 \\ 
\midrule

Italian    & Sanguinetti et al. \cite{SanguinettiEtAlLREC2018} &  1,291 & 5,637 & 6,928   \\ 
\midrule
Total    & & 11,975 &  26,199  &38,174 \\

\bottomrule
\end{tabular}
\end{table}

%% file: bilingual_lexicon.tex
\begin{table}
\caption{Equivalences of the word ``pussy'' in the specific hateful lexicon Hurtlex,  and the generic dictionary MUSE.}\label{tab:bilingual_lexicon}

\begin{tabular}{c c c}
&\Small{Hurtlex} & \Small{MUSE Dictionary}\\ 
\cmidrule(lr){2-2}\cmidrule(lr){3-3}	

\Small{Spanish}  &\parbox{8em}{\centering\Small conejo, concha, chucha, coño, chucha, almeja, punta, vagina, chocha, chocho, chichi, raja} & \parbox{8em}{\centering\Small coño, chocho}       \\
\midrule

\Small{Italian}  &   \parbox{8em}{\centering\Small figa, fregna, potta, gnocca, mona, sorca, patonza, ciciotta, passera, fica, conno, paffia, pincia} & \parbox{8em}{\centering\Small fica, figa}\\
\bottomrule
\end{tabular}
\end{table}

%% file: monolingual_summary.tex
\begin{table*}
\caption{Monolingual hate speech evaluation using F-score. The "Fine-Tuned-Bert" column shows the F-score obtained by fine-tuning monolingual pre-trained Bert. Each of the other columns represents a different word-embedding representations. The best performing models are annotated in the corresponding cell. The numbers on bold letters represent the best performance per training-testing setup.}

\label{tab:monolingual_summary}
\begin{tabular}{@{}llcccccc@{}} %\toprule
&  &  & \multicolumn{4}{c}{Input Representations}\\

\cmidrule(lr){4-7}	

Testing	& Training& \Small{Fine-tuned BERT} &	BERT & mBERT	& MUSE	&HateEmb\\ 
\cmidrule(lr){1-1}\cmidrule(lr){2-2}\cmidrule(lr){3-3}\cmidrule(lr){4-6}\cmidrule(lr){7-7}

 English\_test & English\_train & &\Tiny{LSTMATTN}  & \Tiny{LSTMCNN}&\Tiny{CNN} &\Tiny{LSTM}  & \\
& & 73.45 &\textbf{74.51} &74.40 &70.26 &73.59 & \\

\midrule
Spanish\_test & Spanish\_train && \Tiny{CNN} &\Tiny{CNN} & \Tiny{FNN}& \Tiny{CNN} \\
& & \textbf{73.45 }&70.74&70.15 &67.57 & 69.82& \\

\midrule
Italian\_test & Italian\_train  & & \Tiny{LSTMCNN}&\Tiny{LSTMCNN} &\Tiny{LSTMATTN} & \Tiny{LSTMATTN}& \\
&  &69.37 & \textbf{69.74} & 68.98 & 63.95 & 69.38 & \\
\bottomrule
\end{tabular}
\end{table*}

%% file: Cross_lingual_summary.tex
\begin{table*}
\caption{Cross-lingual hate speech evaluation using F-score.  The representations considered were: representations from multilingual Bert (mBert), MUSE, LASER and our specific hate representations (HateEmb). The numbers on bold letters represent the best performance per training-testing setup.}
\label{tab:Cross_lingual-DNN}
\begin{tabular}{@{}llccccl@{}} %\toprule
&  &  \multicolumn{4}{c}{Input Representations}\\
\cmidrule(lr){3-6}	

Testing	& Training& 	 mBERT	& MUSE	&LASER & HateEmb & (diff)\\ 

\cmidrule(lr){1-1}\cmidrule(lr){2-2}\cmidrule(lr){3-5}\cmidrule(lr){6-6}\cmidrule(lr){7-7}

Spanish\_test & English\_train &\Tiny{LSTMATTN} &\Tiny{LSTM}  &\Tiny{RF} &\Tiny{LSTMCNN}& \\
 &&56.80 &50.73& 52.25&\Large{\textbf{60.64}} & \Small{+0.34} \\

\cmidrule(lr){2-7}

 & Italian\_train&\Tiny{ATTN}  &\Tiny{LSTM}&\Tiny{XGB}& \Tiny{LSTM} \\
& &\Large{\textbf{55.21}} &46.22& 54.70&54.82&  \Small{+0.20}\\

% \cmidrule(lr){-}\cmidrule(lr){-}
\midrule
 Italian\_test & English\_train  & \Tiny{FNN} &\Tiny{LSTM} &\Tiny{XGB} &\Tiny{LSTMATTN} & \\
&&60.73 &55.40& 54.87 & \Large{\textbf{61.82}}&\Small{+0.09} \\

\cmidrule(lr){2-7}

&Spanish\_train&\Tiny{ATTN} & \Tiny{LSTMATTN} &\Tiny{XGB} &\Tiny{LSTM}&\Tiny{}\\
& &\Large{\textbf{62.24}}& 53.81&55.09&58.38& \Small{-3.86}\\
\midrule
 English\_test & Spanish\_train&\Tiny{LSTM}& \Tiny{ATTN} &\Tiny{XGB}&\Tiny{LSTM} & \\
&&61.64&48.19& 53.76&\Large{\textbf{63.91}} & \Small{+1.27}\\

\cmidrule(lr){2-7}

 & Italian\_train&\Tiny{MATTN} & \Tiny{LSTM}&\Tiny{XGB}&\Tiny{CNN}  & \\
&&56.80&43.29 &51.68&\Large{\textbf{58.04}}& \Small{+0.24}\\
\bottomrule
\end{tabular}
\end{table*}
% \input{insert_monolingual}

%% file: neighbors_in_vectors.tex
\begin{table*}
\caption{Terms extracted from the datasets and their Nearest Neighbors in vector spaces of different languages.}
\label{tab:neighbors_in_vectors}

\begin{tabular}{ c c c c c }

 & \multicolumn{2}{c}{\Small{NN in HateEmb space}} & \multicolumn{2}{c}{\Small{NN in Muse space}}\\ 

\cmidrule(lr){2-3}\cmidrule(lr){4-5}

 \Small{Italian Source Term} & \Small{English}  & \Small{Spanish}& \Small{English}  & \Small{Spanish}\\
\cmidrule(lr){1-1}\cmidrule(lr){2-2}\cmidrule(lr){3-3}\cmidrule(lr){4-4}\cmidrule(lr){5-5}
\Small{migranti} &  \Small{illegals}   &  \Small{palestinos}&\Small{migrants}&\Small{migrantes}\\ 

% \Small{terroristi} &  \Small{cockroaches}   &  \Small{protestantes}&\Small{terroristas}&\Small{terroristas}\\ 

% \Small{islamici} &  \Small{islamic}   &  \Small{sionista}&\Small{islamic}&\Small{islamicos}\\

\midrule

\Small{Spanish Source Term} & \Small{English}  & \Small{Italian}& \Small{English}  & \Small{Italian}\\
\cmidrule(lr){1-1}\cmidrule(lr){2-2}\cmidrule(lr){3-3}\cmidrule(lr){4-4}\cmidrule(lr){5-5}

\Small{gitano}&\Small{portuguese}&\Small{negro}&\Small{gypsy}&\Small{gitano} \\
%\cmidrule(lr){2-8}

% \Small{fachas}  & \Small{bigots} &  \Small{comunisti}&---&---\\ 
% % \cmidrule(lr){2-7}

% \Small{negros}&\Small{blacks}&\Small{neri}&\Small{blacks}&\Small{neri} \\ 
 
 %&     &    &    &  &    &     \\ 

\midrule
\Small{English Source Term} & \Small{Spanish}  & \Small{Italian}& \Small{Spanish}  & \Small{Italian}\\

\cmidrule(lr){1-1}\cmidrule(lr){2-2}\cmidrule(lr){3-3}\cmidrule(lr){4-4}\cmidrule(lr){5-5}

% \Small{retarded} & \Small{imbecil}
%     &  \Small{idiota}  & \Small{mental} & \Small{ritardato}\\ 
% %\cmidrule(lr){2-8}

 % \parbox{7em}{\centering\Tiny}               
\Small{muslims} &\Small{musulmanes}   &   \Small{musulmani}&\Small{musulmanes}   &   \Small{musulmani}\\

%\cmidrule(lr){2-8}
% \Small{nicca} &  \parbox{7em}{\centering\Tiny }    & \parbox{7em}{\centering\Tiny }   & \parbox{7em}{\centering\Tiny  mongolo, 0.58,\\ tipo, 0.58,\\ gordo, 0.57}   & \parbox{7em}{\centering\Tiny merda, 0.56,\\ vogliono, 0.56,\\ islam , 0.56} &  \parbox{7em}{\centering\Tiny }  &\parbox{7em}{\centering\Tiny }   \\
% \cmidrule(lr){2-7}
%   \Small{christians} &
%   \Small{catolicos}   &
%   \Small{cattolici}& \Small{cristianos}& \Small{cristiani} \\ 

% &     &    &    &  &    &   \\ 
\bottomrule

\end{tabular}
\end{table*}

%% file: neighbors_in_data.tex
\begin{table*}[!h]
\caption{For some seed terms we found the top most similar terms in the different datasets. The similarities were calculated separately for Hateful and Non-Hateful classes so we can observe different relations depending on the nature of  the expressions. The numbers represent the similarity achieved on each case. As vector representations were use HateEmb.}
\label{tab:Cualitative_evaluation}

\begin{tabular}{ c  c c c c }

 & \multicolumn{2}{c}{\Small{Hateful Data}}  & \multicolumn{2}{c}{\Small{No-Hateful Data}} \\ 

\cmidrule(lr){2-3}\cmidrule(lr){4-5}
\Small{English Seed Term} &\Small{Spanish} & \Small{Italian} & \Small{Spanish} & \Small{Italian}\\

\cmidrule(lr){1-1}\cmidrule(lr){2-2}\cmidrule(lr){3-3}\cmidrule(lr){4-4}\cmidrule(lr){5-5}

 % \parbox{7em}{\centering\Tiny}               
% \Small{latinos}  & \parbox{7em}{\centering\Small civiles --- 0.48 \\guardias --- 0.48\\agresión --- 0.47}

% & \parbox{7em}{\centering\Small sussidi --- 0.61\\ attacchi --- 0.60\\ pedofili --- 0.60} & 
% ---
% & --- \\ 
% \addlinespace

\Small{girls} & \parbox{7em}{\centering\Small perra --- 0.73\\ igual --- 0.73}   & \parbox{7em}{\centering\Small immigrati --- 0.73\\fatto --- 0.71}   &\parbox{7em}{\centering\Small cosas --- 0.69\\ subnormal --- 0.68}  &\parbox{7em}{\centering\Small dire --- 0.68\\arriva --- 0.68} \\ 
%\cmidrule(lr){2-8}

\addlinespace
\Small{muslims} &  \parbox{7em}{\centering\Small putos --- 0.72 \\ fascistas --- 0.71 }
& \parbox{7em}{\centering\Small islamica --- 0.74 \\ italia --- 0.72} & 

\parbox{7em}{\centering\Small migrantes --- 0.70\\guerra --- 0.69}  & 
\parbox{7em}{\centering\Small cristiani --- 0.69\\europa --- 0.69}\\

\midrule

\Small{Spanish Seed Term} & \Small{English} & \Small{Italian} & \Small{English} & \Small{Italian} \\
\cmidrule(lr){1-1}\cmidrule(lr){2-2}\cmidrule(lr){3-3}\cmidrule(lr){4-4}\cmidrule(lr){5-5}
\Small{gitano}& \parbox{7em}{\centering\Small invaders --- 0.66  \\ mexican --- 0.65}    & \parbox{7em}{\centering\Small vivere --- 0.74\\male --- 0.73} &  \parbox{7em}{\centering\Small fingers --- 0.61\\ loving --- 0.61}
 &  \parbox{7em}{\centering\Small caccia --- 0.74\\guida --- 0.74} \\ 
 \addlinespace
 
\Small{palestinos}& \parbox{7em}{\centering\Small disaster --- 0.64\\genocide --- 0.62}    & \parbox{7em}{\centering\Small cancro --- 0.79\\ pericolo --- 0.78} &  \parbox{7em}{\centering\Small iraqi --- 0.64\\fleeing --- 0.64}
 &  \parbox{7em}{\centering\Small 'l’accordo', 0.73\\mette --- 0.73}\\
%\cmidrule(lr){2-8}

% \Small{islam}  & \parbox{7em}{\centering\Small european --- 0.66\\  nations --- 0.65\\ protest --- 0.65}    &\parbox{7em}{\centering\Small civile --- 0.76\\ oriana --- 0.76\\radici --- 0.76}  &  \parbox{7em}{\centering\Small save --- 0.67\\united --- 0.66\\ former --- 0.65}  & \parbox{7em}{\centering\Small povero --- 0.74\\ piccolo --- 0.72\\mille --- 0.72}   \\ 
% % \cmidrule(lr){2-7}
% \addlinespace
 %&     &    &    &  &    &     \\ 

\midrule
 \Small{Italian Seed Term}  & \Small{English} & \Small{Spanish} & \Small{English} & \Small{Spanish} \\
\cmidrule(lr){1-1}\cmidrule(lr){2-2}\cmidrule(lr){3-3}\cmidrule(lr){4-4}\cmidrule(lr){5-5}
\Small{migranti} & \parbox{7em}{\centering\Small living --- 0.74 \\ refugees --- 0.72} &  \parbox{7em}{\centering\Small sudacas --- 0.82 \\ podemos --- 0.82 } &  \parbox{7em}{\centering\Small actually --- 0.72\\something --- 0.71}  & \parbox{7em}{\centering\Small cosas --- 0.82 \\ subnormal --- 0.81}\\ 
\addlinespace
\Small{terroristi} &  \parbox{7em}{\centering\Small muslims --- 0.70 \\ welcome --- 0.69}
& \parbox{7em}{\centering\Small fascistas --- 0.80 \\ putos --- 0.79}  &   \parbox{7em}{\centering\Small actually  --- 0.72 \\ without --- 0.72}  &  \parbox{7em}{\centering\Small ridículo --- 0.82 \\ cosas --- 0.82}\\ 
% \Small{neri} &  \parbox{7em}{\centering\Small terrorists --- 0.65\\possible --- 0.65\\ service --- 0.65}  & \parbox{7em}{\centering\Small africana --- 0.75 \\ puedan --- 0.74 \\ viola --- 0.74}  & \parbox{7em}{\centering\Small without --- 0.70 \\ immigration --- 0.69\\million --- 0.69}   &  \parbox{7em}{\centering\Small sesión --- 0.81 \\habláis --- 0.79\\hoja --- 0.79}  \\
\addlinespace
% &     &    &    &  &    &   \\ 
\bottomrule

\end{tabular}
\end{table*}

%% file: neighbors_in_data_MUSE.tex
\begin{table*}
\caption{Results for an experiment similar to the one presented in Table~\ref{tab:Cualitative_evaluation} but considering the general purpose MUSE multilingual embeddings instead of our hate-specific embeddings.} 
\label{tab:Cualitative_evaluation_MUSE}

\begin{tabular}{ c  c c c c }

 & \multicolumn{2}{c}{\Small{Hateful Data}}  & \multicolumn{2}{c}{\Small{No-Hateful Data}} \\ 

\cmidrule(lr){2-3}\cmidrule(lr){4-5}
\Small{English Seed Term} &\Small{Spanish} & \Small{Italian} & \Small{Spanish} & \Small{Italian}\\

\cmidrule(lr){1-1}\cmidrule(lr){2-2}\cmidrule(lr){3-3}\cmidrule(lr){4-4}\cmidrule(lr){5-5}

 % \parbox{7em}{\centering\Tiny}               
%\cmidrule(lr){2-8}
% \Small{nicca} &  \parbox{7em}{\centering\Tiny }    & \parbox{7em}{\centering\Tiny }   & \parbox{7em}{\centering\Tiny  mongolo  ---  0.58\\ tipo  ---  0.58\\ gordo  ---  0.57}   & \parbox{7em}{\centering\Tiny merda  ---  0.56\\ vogliono  ---  0.56\\ islam   ---  0.56} &  \parbox{7em}{\centering\Tiny }  &\parbox{7em}{\centering\Tiny }   \\
% \cmidrule(lr){2-7}
\Small{girls} & \parbox{7em}{\centering\Small gusta --- 0.75\\ mujeres --- 0.74}   & \parbox{7em}{\centering\Small  islamico --- 0.71 \\ stranieri --- 0.71}   & \parbox{7em}{\centering\Small mamón --- 0.70 \\ eres --- 0.70}&\parbox{7em}{\centering\Small vedere --- 0.69 \\ bene --- 0.68} \\ 
%\cmidrule(lr){2-8}
\addlinespace
\Small{muslims} &  \parbox{7em}{\centering\Small españa --- 0.76 \\ musulmanes --- 0.72}  & \parbox{7em}{\centering\Small ecco --- 0.76\\italia --- 0.75} &  \parbox{7em}{\centering\Small guerra --- 0.73\\ llegan --- 0.71}  & \parbox{7em}{\centering\Small pero --- 0.76\\ europa --- 0.76}\\

\midrule

\Small{Spanish Seed Term} & \Small{English} & \Small{Italian} & \Small{English} & \Small{Italian} \\
\cmidrule(lr){1-1}\cmidrule(lr){2-2}\cmidrule(lr){3-3}\cmidrule(lr){4-4}\cmidrule(lr){5-5}

\Small{gitano}& \parbox{7em}{\centering\Small hopefully --- 0.70\\complete --- 0.70}    & \parbox{7em}{\centering\Small credo --- 0.68\\pensa --- 0.68} &  \parbox{7em}{\centering\Small cock --- 0.68\\ gotta --- 0.67}
 &  \parbox{7em}{\centering\Small cambiare --- 0.66\\ stranieri --- 0.66} \\ 
 \addlinespace

%\cmidrule(lr){2-8}

\Small{palestinos}  & \parbox{7em}{\centering\Small  beaners --- 0.61\\disaster --- 0.61}    &\parbox{7em}{\centering\Small duce --- 0.68 \\ bestie --- 0.67}  &  \parbox{7em}{\centering\Small return --- 0.69 \\ syrian --- 0.68}  & \parbox{7em}{\centering\Small largo --- 0.69 \\ corridoi --- 0.69}   \\ 
% \cmidrule(lr){2-7}
 %&     &    &    &  &    &     \\ 

\midrule
 \Small{Italian Seed Term}  & \Small{English} & \Small{Spanish} & \Small{English} & \Small{Spanish} \\
\cmidrule(lr){1-1}\cmidrule(lr){2-2}\cmidrule(lr){3-3}\cmidrule(lr){4-4}\cmidrule(lr){5-5}

\Small{migranti} &  \parbox{7em}{\centering\Small refugees --- 0.73\\living --- 0.72}   & \parbox{7em}{\centering\Small refugiados --- 0.72\\quiere --- 0.70}  &   \parbox{7em}{\centering\Small migrants --- 0.74 \\ take --- 0.71}  &  \parbox{7em}{\centering\Small mamón --- 0.74\\catalán --- 0.71}   \\ 
\addlinespace

\Small{terroristi} &  \parbox{7em}{\centering\Small religion --- 0.75 \\ quran --- 0.75}  & \parbox{7em}{\centering\Small malditos --- 0.73\\mientras --- 0.724}  & \parbox{7em}{\centering\Small migrants --- 0.73 \\ready --- 0.72}   &  \parbox{7em}{\centering\Small viendo --- 0.72 \\mamón --- 0.72}  \\
% &     &    &    &  &    &   \\ 
\bottomrule

\end{tabular}
\end{table*}